\documentclass[conference]{IEEEtran}
\IEEEoverridecommandlockouts
\usepackage{cite}
\usepackage{amsmath,amssymb,amsfonts}
\usepackage{graphicx}
\usepackage{textcomp}
\usepackage{xcolor}
\def\BibTeX{{\rm B\kern-.05em{\sc i\kern-.025em b}\kern-.08em
    T\kern-.1667em\lower.7ex\hbox{E}\kern-.125emX}}
    
\usepackage{algorithm}
\usepackage{algpseudocode}
\newcommand{\vars}{\texttt}
\newcommand{\func}{\textrm}
\let\oldReturn\Return
\renewcommand{\Return}{\State\oldReturn}

\usepackage{hyperref}

\usepackage{subcaption}

\usepackage{multirow}

\begin{document}

\title{Real-Time Kinodynamic Motion Planning for Omnidirectional Mobile Robot Soccer using Rapidly-Exploring Random Tree in Dynamic Environment with Moving Obstacles\\
}

\author{\IEEEauthorblockN{Fahri Ali Rahman$^{1}$\thanks{\emph{$^{1}$Student, Department of Electrical Engineering and Information Technology, Faculty of Engineering Universitas Gadjah Mada, Yogyakarta, Indonesia; email: fahri.ali.r@mail.ugm.ac.id}}, Igi Ardiyanto$^{2}$\thanks{\emph{$^{2,3}$Department of Electrical Engineering and Information Technology, Faculty of Engineering Universitas Gadjah Mada, Yogyakarta, Indonesia; email: \{igi; adha.imam\}@ugm.ac.id}}, Adha Imam Cahyadi$^{3}$}
%\IEEEauthorblockA{\textit{Department of Electrical Engineering and Information Technology} \\
%\textit{Faculty of Engineering Universitas Gadjah Mada}\\ Yogyakarta, Indonesia \\ fahri.ali.r@mail.ugm.ac.id, \{igi; adha.imam\}@ugm.ac.id}
%\and
%\and
}

\maketitle

\begin{abstract}
RoboCup Middle Size League (RoboCup MSL) provides a standardized testbed for research on mobile robot navigation, multi-robot cooperation, communication and integration via robot soccer competition in which the environment is highly dynamic and adversarial. One of important research topic in such area is kinodynamic motion planning that plan the trajectory of the robot while avoiding obstacles and obeying its dynamics. 
Kinodynamic motion planning for omnidirectional robot based on kinodynamic-RRT* method is presented in this work. Trajectory tracking control to execute the planned trajectory is also considered in this work. Robot motion planning in translational and rotational direction are decoupled. Then we implemented kinodynamic-RRT* with double integrator model to plan the translational trajectory. The rotational trajectory is generated using minimum-time trajectory generator satisfying velocity and acceleration constraints. The planned trajectory is then tracked using PI-Control. To address changing environment, we developed concurrent sofware module for motion planning and trajectory tracking. 
The resulting system were applied and tested using RoboCup simulation system based on Robot Operating System (ROS). The simulation results that the motion planning system are able to generate collision-free trajectory and the trajectory tracking system are able to follow the generated trajectory. It is also shown that in highly dynamic environment the online scheme are able to re-plan the trajectory.
\end{abstract}

\vspace{5mm} %5mm vertical space

\begin{IEEEkeywords}
Kinodynamic Motion Planning, Omnidirectional Mobile Robot, Dynamic Environment, Trajectory Tracking, Online Planning
\end{IEEEkeywords}

\section{Introduction}

RoboCup is an international robot competition that attempting to foster AI and intelligent robotics research \cite{kitano1997robocup}. By providing common problem and evaluation where a wide range of technologies need to be integrated, it includes multiple areas of robotic research and AI, such as multi-agent collaboration, real-time reasoning and planning, intelligent robotics and sensor fusion. RoboCup Middle-Size League (RoboCup MSL) is a robot soccer competition of two teams with $50 x 50 x 80$-cm sized robots playing with each other \cite{burkhard2002road}. Where the robots are fully autonomous and required sensors and power are on-board. Such competition settings leads to a very dynamic and adversarial environment \cite{burkhard2002road}. In case of soccer robot, it is also desirable for the robot to be able move safely from one point to another while colliding with obstacles, hence, it is necessary to solve the motion planning problem. Furthermore, in such dynamic environment where the obstacles are obviously not a static object, it is preferred to anticipate the movement of the obstacles while solving the planning problem. In addition, when the robot moving in high speed, it becomes necessary to address the dynamics of the robot. Therefore, this work aimed to resolve the kinodynamic motion planning problem for omnidirectional mobile robot with moving obstacles in robot soccer context.

In \cite{manalu2014double}, Manalu developed movement planning in case of passing and kicking the ball using Double Target Potential Field while assuming the obstacles as static object. Afakh et. al. implemented fuzzy logic controller to  optimize bicycle path planning on omnidirectional mobile robot \cite{afakh2018bicycle}. In \cite{albab2017path}, Albab et. al. generate paths for omnidirectional robot soccer that avoid some static obstacles using genetic algorithm that will generate random via points and encode the paths as gene. It is reported that the system is able to avoid obstacles in static environment, and it is also tested on dynamic environment by updating the path every time the environment changing but still assuming static obstacles. Dikairono et. al. presented motion planning simulator and implemented A* path planning algorithm for obstacle avoidance while treating the obstacles as a static object \cite{dikairono2017motion}. The simulator is based only on robot's kinematics. The experiments shows that the robot successfully create a path that avoid the obstacles. However, it is unclear whether the proposed system are successfully avoid moving obstacles. In this work, we also considered the dynamics of the robot and explicitly model the movement of the obstacles.

The remainder of this paper is organized as follows. In section \ref{modeling}, we discuss the kinematic model of omnidirectional robot, environment model, and double integrator dynamics. In section \ref{planning_tracking}, we discuss the proposed motion planning and trajectory tracking system. In section \ref{result}, we report our numerical and dynamic simulation. Finally, we conclude in section \ref{conclusion}.

\section{Robot and Environment Model} \label{modeling}

\subsection{Kinematics of Omnidirectional Robot}

An omnidirectional robot belongs to holonomic robot family. It can move in translational and rotational space simultaneously and independently. The kinematic equation of the robot is given by 2-dimensional coordinate transformation in from the body frame to the world frame :
%\begin{figure}
%	\centering
%	\includegraphics[width=0.2\textwidth]{omni-frame}
%	\caption{Illustration of fixed World Frame and Robot's Body Frame}
%	\label{frame}
%\end{figure}
\begin{equation} \label{omnirobot_eq}
\begin{bmatrix}
u\\
v\\
r
\end{bmatrix}
=
\begin{bmatrix}
\cos{\omega(t)} & \sin{\omega(t)} & 0 \\
-\sin{\omega(t)} & \cos{\omega(t)} & 0 \\
0 & 0 & 1
\end{bmatrix}
\begin{bmatrix}\dot{x}
\\ \dot{y}
\\ \dot{\omega}
\end{bmatrix}
\end{equation}
where $\dot{x}$ and $\dot{y}$ are the translational velocity in $x$ and $y$ direction of the robot in the world frame, respectively. While $u$ and $v$ are the translational velocity of the robot in $x$ and $y$ direction of the robot in the body frame, respectively. It should be noted that from the transformation matrix above, the rotational velocity in the world frame, $\dot{\omega}$, equals to the rotational velocity in the robot's body frame, $r$. Moreover, we could maps the velocity in the world frame to velocity in the body frame by inverting the transformation matrix which is invertible for every $\omega$ values.

\subsection{Geometry Simplification of Robot and Obstacles}

In this subsection, we describe our strategy for simplyfing the geomoetry of the robot and obstacles. In order to simplify the collision checking algorithm needed by kinodynamic-RRT*, we simplify the robot as a point and the obstacles as circular objects.

\begin{figure}[ht!]
	\centering
	\includegraphics[width=0.35\textwidth]{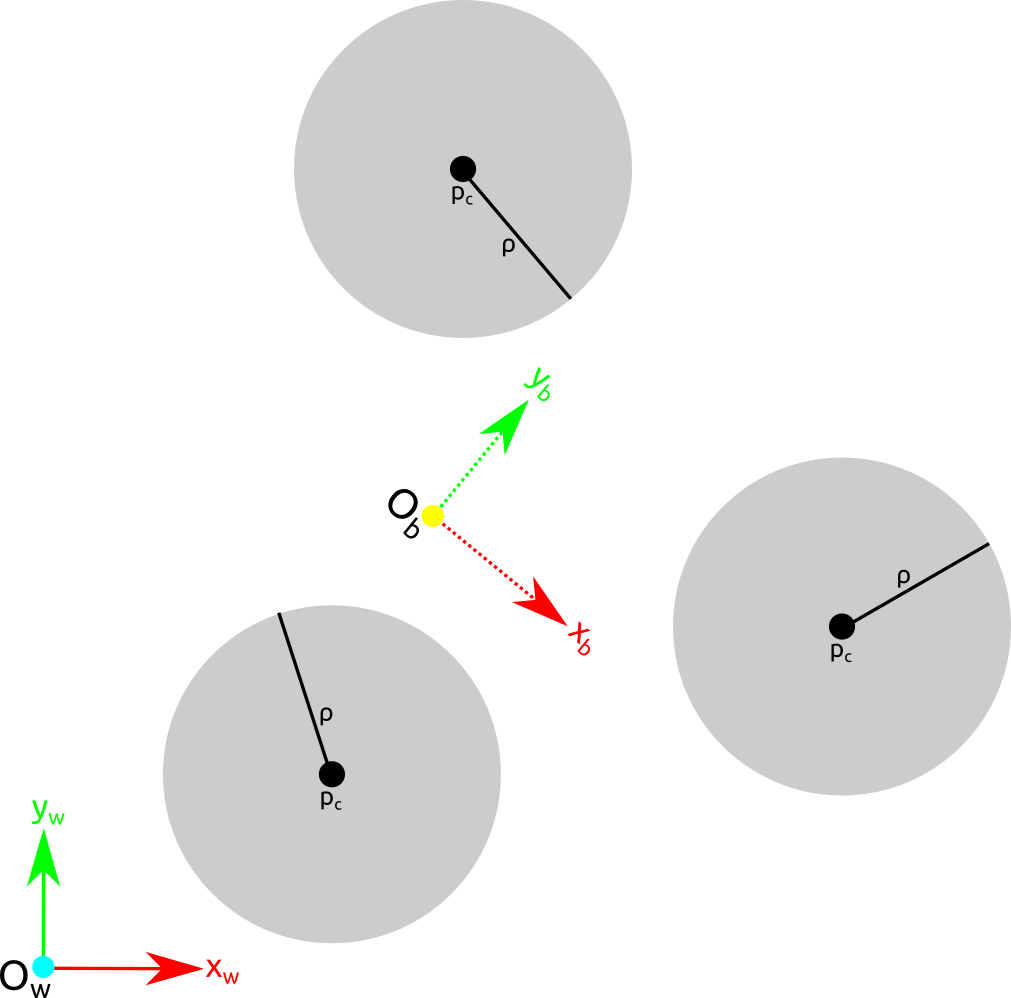}
	\caption{Geometry Simplification of the robot and Obstacles}
	\label{geom_simple}
\end{figure}

Figure \ref{geom_simple} shows the illustration of this simplification. In the figure, the robot is shown as a yellow point with dashed line in red and yellow represent its orientation in the world frame which is shown as blue point. The obstacles are depicted in gray circles with its center point position and radius are ilustrated with $p_0$ and $\rho$, respectively. By simplifying the robot's geometry as a point, the rotational motion of the robot does not contribute to the collision checking.

\subsection{Dynamic Model of the Robot}

Having a simplified geometric model of the robot, we discussed the dynamic model of the robot in this section. In this work, we considered the \emph{NuBot} robot model from Yu et al. in \cite{yu2010nubot}. The robot is treated as \emph{point-mass} object obeying Newton's Law of Motion. Since the robot are able to move in any translational and rotational simultaneusly, we also decoupled the robot's motion in translational and rotational direction. We also assume a \emph{frictionless} ground plane for the purpose of motion planning. The motion in $x$ and $y$ direction as well as rotational direction, $\omega$, are assumed to be independent from each other, so that :

\begin{equation}
\ddot{x} = \frac{1}{m}f_x,\; \ddot{y} = \frac{1}{m}f_y,\; \ddot{\omega} = \frac{1}{m}f_{\omega}
\end{equation}
where $f_x$ and $f_y,$ are the force acting in each direction. While $f_{\omega}$ is the torque.

\subsection{Dynamic Environment with Moving Obstacles}

This work considered the environment which robot operate is a dynamic environment with soccer robot competition setting. To be able plan effectively in the presence of moving obstacles, it is desirable to take the motion of the obstacles while planning the motion. In this Work, we used 3D time-space representation, as defined by Ardiyanto and Miura in \cite{ardiyanto2012real}. The position of the moving obstacles, at time $t$, with constant velocity could be expressed as :
\begin{equation} \label{dynamic_obs_x}
 D^{'}_{x}(t) = D^{'}_{x}(0) + v_{D_x}t
\end{equation}
\begin{equation} \label{dynamic_obs_y}
 D^{'}_{y}(t) = D^{'}_{y}(0) + v_{D_y}t
\end{equation}
where $D^{'}_{x}(t)$ and $D^{'}_{y}(t)$ represents the predicted position of the obstacles at time $t$ in $x$ and $y$ axis of world frame, respectively, while $D^{'}_{x}(0)$ and $D^{'}_{y}(0)$ are the current position of the obstacles and $v_{D_x}$ and $v_{D_y}$ are the velocity of the obstacles.

\subsection{Fixed-Final-State Free-Final-Time Control for Double Integrator}

This subsection describe the derivation of fixed-state free-final-time controller to solve the trajectory between two states obeying double integrator dynamics. We derive the controller from fixed-final-state free-final-time for general system described in \cite{webb2013kinodynamic}.
The state-space equation for double integrator model is given by:
\begin{equation} \label{dbl_int_state_space_eq}
\dot{\boldsymbol{x}} = \begin{bmatrix}
0 & I \\
0 & 0
\end{bmatrix}\boldsymbol{x} + \begin{bmatrix}
0 \\
I
\end{bmatrix}\boldsymbol{u}
\end{equation}
where, 
\begin{equation*} \label{dbl_int_label_eq}
\boldsymbol{x} = 
\begin{bmatrix}
\boldsymbol{p} &
\boldsymbol{v}
\end{bmatrix}^{T},
\boldsymbol{u} = \boldsymbol{a},
\end{equation*}
\begin{equation*}
\boldsymbol{p} = 
\begin{bmatrix} 
p_x &
p_y
\end{bmatrix}^{T},
\boldsymbol{v} =
\begin{bmatrix}
v_x &
v_y
\end{bmatrix}^{T},
\boldsymbol{a} = 
\begin{bmatrix}
a_x &
a_y
\end{bmatrix}^{T}
\end{equation*}
where $p_x, p_y, v_x, v_y, a_x,$ and $a_y$ are the position, velocity, and acceleration of the robot in the $x$, and $y$ direction in the world frame, respectively. The trajectory of the robot is defined by a tuple $\pi = (\boldsymbol{x}[], \boldsymbol{u}[], \tau)$. The cost of a trajectory $\pi$ is defined as :
\begin{equation} \label{cost_eq}
c(\pi) = \int_{0}^{\tau}(1+\boldsymbol{u}(t)^{T}R\boldsymbol{u}(t))dt, \quad R = rI
\end{equation}
where $r$ is the input weighting constant.

As proposed in \cite{webb2013kinodynamic}, to solve the optimal trajectory, it is required to solve the \emph{weighted controllability Gramian} and the optimal cost of the double integrator model. The \emph{weighted controllability Gramian} for double integrator dynamics id defined as :
\begin{equation} \label{dbl_int_gramian_eq}
G(t)=\begin{bmatrix}
\frac{t^{3}}{3r} & 0 & \frac{t^{2}}{2r} & 0 \\
0 & \frac{t^{3}}{3r} & 0 & \frac{t^{2}}{2r} \\
\frac{t^{2}}{2r} & 0 & \frac{t}{r} & 0\\
0 & \frac{t^{2}}{2r} & 0 & \frac{t}{r}
\end{bmatrix}
\end{equation}
and the cost for double integrator dynamics :
\begin{equation} \label{dbl_int_cost_eq}
c(\tau) =
\tau +
\left \{  
\boldsymbol{x}_f -
\bar{\boldsymbol{x}}(\tau)
\boldsymbol{x}_i
\right \}^{T}
G^{-1}(\tau)
\left \{  
\boldsymbol{x}_f -
\bar{\boldsymbol{x}}(\tau)
\boldsymbol{x}_i
\right \}
\end{equation}
where
\begin{equation} \label{ext_dbl_int_eq}
\bar{\boldsymbol{x}}(t) =
e^{At} \boldsymbol{x}_i 
%= \begin{bmatrix}
%1 & 0 & t & 0 \\
%0 & 1 & 0 & t \\
%0 & 0 & 1 & 0 \\
%0 & 0 & 0 & 1
%\end{bmatrix}
%\boldsymbol{x}_i
\end{equation}
\begin{equation} \label{d_sym_dbl_int_eq}
\boldsymbol{d}(\tau) =
G^{-1}(\tau)
\left \{
\boldsymbol{x}_f -
\bar{\boldsymbol{x}}(\tau)
\boldsymbol{x}_i \right \},
\end{equation}
%define $\boldsymbol{x}_f = \begin{bmatrix}
%p_{xf} & p_{yf} & v_{xf} & v_{yf}
%\end{bmatrix}^{T}$ and $\boldsymbol{x}_i = \begin{bmatrix}
%p_{xi} & p_{yi} & v_{xi} & v_{yi}
%\end{bmatrix}^{T}$, we could expand the equation \ref{d_sym_dbl_int_eq} as :
%\begin{equation*}
%\boldsymbol{d}(\tau) =
%\begin{bmatrix}
%\frac{12r}{t^{3}} & 0 & \frac{-6r}{t^{2}} & 0 \\
%0 & \frac{12r}{t^{3}} & 0 & \frac{-6r}{t^{2}} \\
%\frac{-6r}{t^{2}} & 0 & \frac{4r}{t} & 0 \\
%0 & \frac{-6r}{t^{2}} & 0 & \frac{4r}{t}
%\end{bmatrix}
%\begin{bmatrix}
%p_{xf} - p_{xi} - v_{xi}t\\
%p_{yf} - p_{yi} - v_{yi}t\\
%v_{xf} - v_{xi}\\
%v_{yf} - v_{yi}
%\end{bmatrix},
%\end{equation*}
%\begin{equation*}
%\boldsymbol{d}(\tau) =
%\begin{bmatrix}
%\frac{-6r(v_{xf}-v_{xi})}{t^2} + \frac{12r(p_{xf}-p_{xi}-v_{xi}t)}{t^3}\\
%\frac{-6r(v_{yf}-v_{yi})}{t^2} + \frac{12r(p_{yf}-p_{yi}-v_{yi}t)}{t^3} \\
%\frac{4r(v_{xf}-v_{xi})}{t} - \frac{6r(p_{xf}-p_{xi}-v_{xi}t)}{t^2}\\
%\frac{4r(v_{yf}-v_{yi})}{t} - \frac{6r(p_{yf}-p_{yi}-v_{yi}t)}{t^2}
%\end{bmatrix}
%\end{equation*}
%and we have closed expression of the derivative of the cost :

Webb et al. suggest that the optimal arrival time, $\tau^{*}$, for the trajectory is at the time when the derivative of \eqref{dbl_int_cost_eq} equals zero \cite{webb2013kinodynamic}. The derivative of \eqref{dbl_int_cost_eq} is given by : 

\begin{equation} \label{dbl_int_cost_derivative_eq}
\dot{c}(\tau) = 1-2\left \{ \begin{bmatrix}
0 & I \\
0 & 0 
\end{bmatrix}
\boldsymbol{x}_f
\right \}^{T}\boldsymbol{d}(\tau)-
\boldsymbol{d}(\tau)^{T}
\begin{bmatrix}
0 & 0 \\
0 & R^{-1}
\end{bmatrix}
\boldsymbol{d}(\tau),
\end{equation}
in this work, to find the root of \eqref{dbl_int_cost_derivative_eq}, we use secant method \cite{chapra1998numerical}. Furthermore, since the time is needed to be positive value, we simply restart the search of the root if it finds negative value. Finally, the optimal trajectory is given by:
\begin{equation} \label{dbl_int_composite_state_eq}
\begin{bmatrix}
\boldsymbol{x}(t)\\
\boldsymbol{y}(t)
\end{bmatrix}
=
\begin{bmatrix}
I &
A_1 & A_2 & A_3
\\
0 & I & A_4 & A_5 \\
0 & 0 & I & 0 \\
0 & 0 & A_6 & I
\end{bmatrix}
\begin{bmatrix}
\boldsymbol{x}_f\\
\boldsymbol{d}(\tau^{*})
\end{bmatrix}
\end{equation}
where 
%\begin{equation} \label{y_sym_eq}
%\boldsymbol{y}(t) = e^{A^{T}(\tau^{*})}\boldsymbol{d}(\tau^{*})
%\end{equation}
\begin{equation} \label{u_eq}
\boldsymbol{u}(t) = R^{-1}B^{T}\boldsymbol{y}(t)
\end{equation}
\begin{equation*}
A_1 = 
\begin{bmatrix}
t-\tau & 0 \\
0 & t-\tau
\end{bmatrix}
,\quad
A_2 = 
\begin{bmatrix}
\frac{(-t+\tau)(t-\tau)^2}{6r} & 0 \\
0 & \frac{(-t+\tau)(t-\tau)^2}{6r}
\end{bmatrix}
\end{equation*}
\begin{equation*}
A_3 = \begin{bmatrix}
\frac{(t-\tau)^2}{2r} & 0 \\
0 & \frac{(t-\tau)^2}{2r}
\end{bmatrix}
,\quad
A_6 = \begin{bmatrix}
-t+\tau & 0 \\
0 & -t+\tau
\end{bmatrix}
\end{equation*}
\begin{equation*}
A_5 = \begin{bmatrix}
\frac{t-\tau}{r} & 0\\
0 & \frac{t-\tau}{r}
\end{bmatrix}
,\quad
A_4 = \begin{bmatrix}
\frac{(-t+\tau)(t-\tau)}{2r} & 0 \\
0 & \frac{(-t+\tau)(t-\tau)}{2r}
\end{bmatrix} 
\end{equation*}

Equation (\ref{dbl_int_composite_state_eq}) above solves the trajectory connecting initial state $x_i$ to final stat $x_f$ satisfying double integrator dynamics. This trajectory solver is used by the following Kinodynamic RRT* algorithm to determine the connection between state.

\section{Kinodynamic Motion Planning and Trajectory Tracking for Omnidirectional Robot} \label{planning_tracking}

\subsection{Motion Planning Strategy}\label{MPS}
The main idea of the strategy is to decouple the trajectory planning for translational motion and rotational motion, taking advantages of the holonomic property of the omnidirectional mobile robot to simplify the collision checking by model the robot as point mass with double integrator dynamics. This simplification leads to simpler algorithm since only translational motion contribute to the collision while the rotational motion can be solved afterwards. Trajectory planning for the translational motion is done by the kinodynamic-RRT*, and the trajectory planning for the rotational motion is done by minimum-time trajectory generator with velocity and acceleration constraints \cite{purwin2006trajectory}.

\subsection{Kinodynamic-RRT* for Double Integrator Model in Dynamic Environment}

We implemented kinodynamic-RRT* algorithm, as proposed by Webb and Berg in \cite{webb2013kinodynamic}, with double integrator model to deal with dynamic environment. Algorithm \ref{impl_rrt} shows the implemented kinodynamic-RRT* algorithm. Furthermore, we also apply stochastic decision to directly sampling the goal state as proposed in \cite{bruce2002real}, to improve the quality and consistency of solution trajectory.

The algorithm takes initial state, $\boldsymbol{q}_{start}$, goal state, $\boldsymbol{q}$, probability of direct sampling, $p$, and maximum number of iteration $n_{iter}$. Unlike the original kinodynamic-RRT*, we restrict the number of iteration to $n_{iter}$, instead of infinitely growing the tree. 

\begin{algorithm}[ht!]
\begin{algorithmic}[1]
\Function{DoubleIntKinodynamic-RRT*}{$\boldsymbol{q}_{start}, \boldsymbol{q}_{goal}, p, n_{iter}$}
	\State $\mathcal{T} \leftarrow \{\boldsymbol{q}_{start}\}$
	\For{ $i\in [1,n_{iter}]$ }
		\State $\boldsymbol{q}_i \leftarrow \func{Sample}(\boldsymbol{q}_{goal})$
		\State $\boldsymbol{q}_{neighbor} \leftarrow \mathcal{T} \mid c^{*}(\boldsymbol{q},\boldsymbol{q}_i) < r \wedge \, not \,\textsc{TrajectoryCollision}(\pi^{*}(\boldsymbol{q},\boldsymbol{q}_i))$
		\State $\boldsymbol{q} \leftarrow \func{argmin} \left(\{ \boldsymbol{q} \in \boldsymbol{q}_{neighbor} \right\}(c^{*}(\boldsymbol{q},\boldsymbol{q}_i))$
		\State $parent(\boldsymbol{q}_i) \leftarrow \boldsymbol{q}$
		\State $cost(\boldsymbol{q}_i) \leftarrow cost(\boldsymbol{q}) + c^{*}(\boldsymbol{q},\boldsymbol{q}_i)$
		\State
			$\boldsymbol{q}_{neighbor} \leftarrow \mathcal{T} \cup \{q_{goal}\} \mid c^{*}(\boldsymbol{q}_i,\boldsymbol{q}) < r $
			\par\hskip\algorithmicindent $\wedge \, not \, \textsc{TrajectoryCollision}(\pi^{*}(\boldsymbol{q}_i,\boldsymbol{q}))$
			%\end{varwidth}
		\For{$\boldsymbol{q} \in \boldsymbol{q}_{neighbor}$}
			\If{$cost(\boldsymbol{q}_i)+c^{*}(\boldsymbol{q}_i,\boldsymbol{q})<cost(\boldsymbol{q})$}
				\State $parent(\boldsymbol{q}) \leftarrow \boldsymbol{q}_i$
				\State $cost(\boldsymbol{q}) \leftarrow cost(\boldsymbol{q}_i)+c^{*}(\boldsymbol{q}_i,\boldsymbol{q})$
			\EndIf
		\EndFor
		\If{$parent(q_i)$}
			\State $\mathcal{T}\leftarrow\mathcal{T}\cup q_i$
		\EndIf
	\EndFor
	\State $\boldsymbol{q}_{solution}[]\leftarrow\emptyset$
	\State $\boldsymbol{q}\leftarrow\boldsymbol{q}_{goal}$
	\While {$parent(\boldsymbol{q})$}
		\State $\func{insert}\;\pi^{*}(parent(\boldsymbol{q}),\boldsymbol{q})\; \func{to}\; \boldsymbol{q}_{solution}[]$
		\State $\boldsymbol{q}\leftarrow parent(\boldsymbol{q})$
	\EndWhile
	\Return $\boldsymbol{q}_{solution}[]$
\EndFunction
\end{algorithmic}
\caption{Implemented Kinodynamic-RRT* Algorithm}
\label{impl_rrt}
\end{algorithm}

The optimal control policy, $\pi^{*}$, in line $5$ and $9$ implements the fixed-final-state free-final-time control for double integrator model from \eqref{dbl_int_composite_state_eq}. The cost of the optimal control policy, $c^{*}$, needed in line $6$ and $12$, is calculated using the cost of optimal control for double integrator model in \eqref{dbl_int_cost_eq}. We implement collision-checking algorithm, needed by \textsc{TrajectoryCollisionCheck} procedure in line $5$ and line $6$, by approximating the trajectory as sequence of lines and moving obstacles as collection of circles and check the collision by performing collision check for line and circle \cite{bourke2016points}. Line $6$ of the algorithm select the neighboring states of the sampled state that has lower cost among the other.

After the tree is grown, the solution trajectory, $\boldsymbol{q}_{solution}[]$ is returned by backtracking the goal state to initial state in line $17$ to $20$. In this work, we use KD-Tree representation from \cite{blanco2014nanoflann}.

\subsection{Trajectory Planning for Rotational Motion}

We implemented a minimum-time trajectory generation algorithm, as proposed by Purwin et al. in \cite{purwin2006trajectory}, for our rotational trajectory planning. The algorithm generate a 1-dimensional optimal trajectory that minimize the travel time subject to velocity and acceleration limit. The algorithm works by computing control effort, terminal state, and arrival time by applying different case depending on the initial position and velocity, as well as final state and constraints of the system. We refer the readers to \cite{purwin2006trajectory} for more detailed description.

\subsection{Trajectory Tracking for Omnidirectional Mobile Robot}
In this subsection, we discuss the implemented trajectory tracking for omnidirectional mobile robot. In this work, we use 'outer-loop' Trajectory Tracking as proposed in \cite{liu2008omni} to control the robot to follow the planned trajectory as reference. Unlike the original trajectory tracking as proposed in \cite{liu2008omni}, we do not use pseudo-differentiator to approximate the derivative of the desired trajectory since the outputted trajectory from motion planning module obeying the robots dynamic.

The error of trajectory is defined as :
\begin{equation*}
\boldsymbol{e} = 
\begin{bmatrix}
x(t) &
y(t) & 
\omega(t)
\end{bmatrix}^{T} - 
\begin{bmatrix}
x_d(t) &
y_d(t) & 
\omega_d(t)
\end{bmatrix}^{T}
\end{equation*}
where $[x(t), y(t), \omega(t)]^{T}$ is the position of the robot at time $t$, while $[x_d(t),y_d(t),\omega_d(t)]^{T}$ is the position of the desired trajectory at time $t$. %and $\boldsymbol{e}=\left[e_x\, e_y\, e_\omega \right]^T$.

In \cite{liu2008omni}, to stabilize the tracking error, a proportional-integral (PI) feedback control is proposed:

\begin{equation}
\begin{bmatrix}
\tilde{u} & \tilde{v} & \tilde{r}
\end{bmatrix}^{T} = 
-K_{P}\,\boldsymbol{e}
-K_{I}\int \boldsymbol{e}\,dt
\end{equation}
where $K_P$ is a constant for proportional term and $K_I$ is a constant for integral term.

\begin{equation}
K_{I} = -B_{1}^{-1} diag(-a_{I1}, -a_{I2}, -a_{I3})
\end{equation}
\begin{equation}
K_{P} = 
-B_{1}^{-1} 
( A_1 - diag(-a_{P1},-a_{P2},-a_{P3}))
\end{equation}
where
\begin{equation}
A_1(t)=
\begin{bmatrix}
0 & 0 & -u_d(t)sin(\omega_d(t)) - v_d(t)cos(\omega_d(t)) \\ 
0 & 0 & u_d(t)cos(\omega_d(t))-v_d(t)sin(\omega_d(t)) \\ 
0 & 0 & 0
\end{bmatrix}
\end{equation}

\begin{equation}
B_1(t) =
\begin{bmatrix}cos(\omega_d(t)) & -sin(\omega_d(t)) & 0 \\ 
sin(\omega_d(t)) & cos(\omega_d(t)) & 0 \\ 
0 & 0 & 1
\end{bmatrix}
\end{equation}
Finally, the commanded velocity in the body frame is given by :
\begin{equation}
\begin{bmatrix}
u_{com} &
v_{com} &
r_{com} 
\end{bmatrix}^{T}
=
\begin{bmatrix}
u_d &
v_d &
r_d
\end{bmatrix}^{T}
+
\begin{bmatrix}
\tilde{u} &
\tilde{v} &
\tilde{r}
\end{bmatrix}^{T}
\end{equation}

\begin{figure}[ht!]
	\centering
	\includegraphics[width=0.4\textwidth]{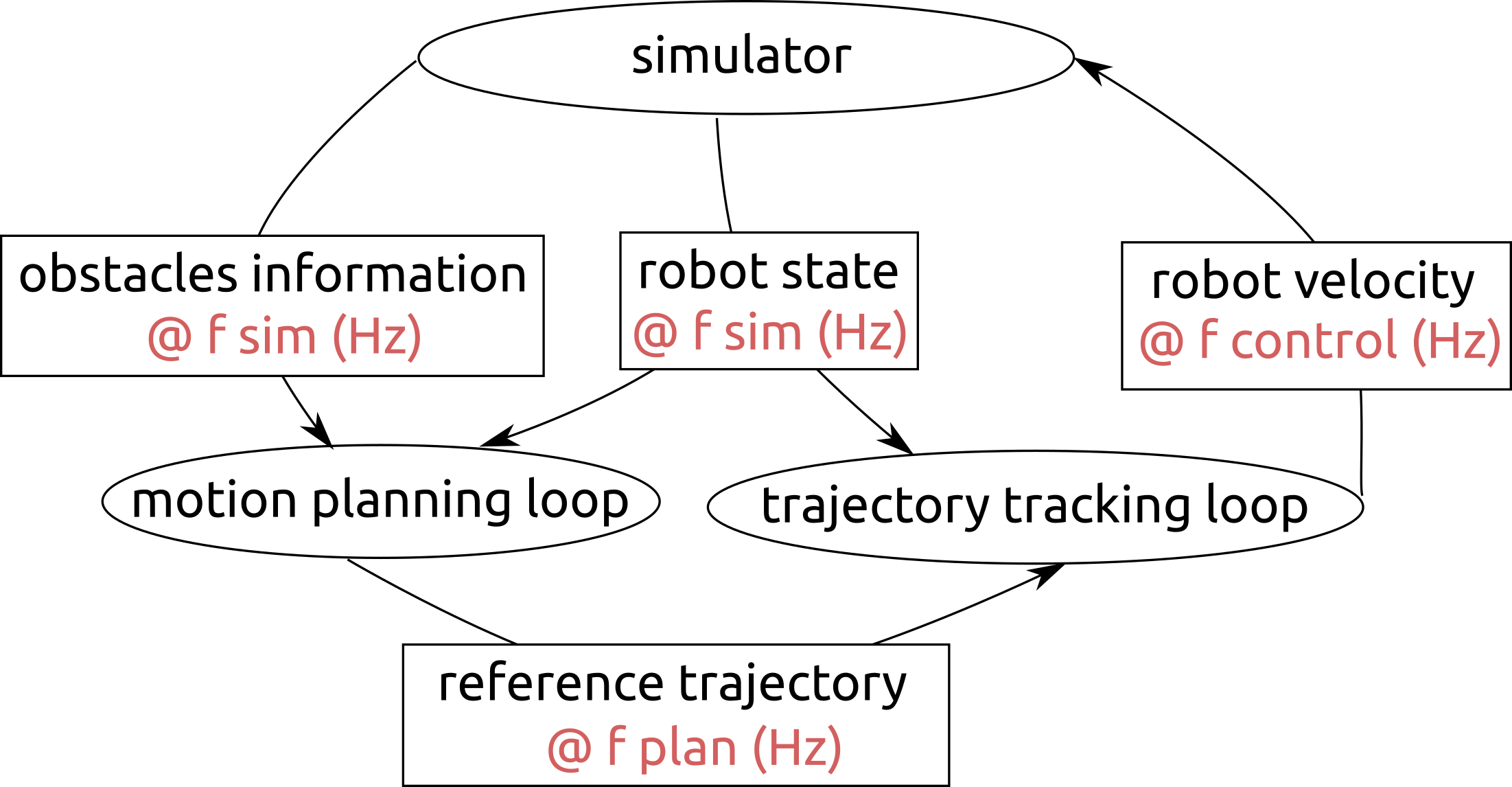}
	\caption{Online Computation of Motion Planning and Trajectory Tracking}
	\label{online_loop}
\end{figure}

\subsection{Online Computation of Motion Planning and Trajectory Tracking}

We tackle the problem of changing environment by executing motion planning and trajectory tracking module in online fashion. Figure \ref{online_loop} illustrates our online computation approach. With this approach, we could compute motion plan and track the trajectory separately at different rate.

In figure \ref{online_loop}, each ellipses illustrate different processes that are executed concurrently and each box represents data that are in the form of ROS messages, with \vars{f sim}, \vars{f plan}, and \vars{f control} represents the loop rate of simulator, motion planning, and trajectory tracking loop, respectively. The simulator process, in this experiment, is the Gazebo simulator from standard ROS distribution.

\section{Experiment Result} \label{result}

We performed several experiment to verify the proposed kinodynamic motion planning system for omnidirectional mobile robot. These Experiment were executed on Laptop PC with Intel Core i3 2.0 GHz processor, 4 GB memory and Ubuntu 16.04 Operating System.

\subsection{Dynamic Constraints Consideration}
\begin{figure}[ht!]
	\centering
	\begin{subfigure}{0.4\textwidth}
		\centering
		\includegraphics[width=1\linewidth]{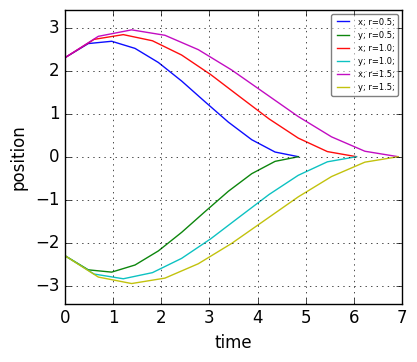}
		\caption{Comparison of position vs. time with different input weight}
		\label{trajectory_pos}
	\end{subfigure}
	\begin{subfigure}{0.4\textwidth}
		\centering
		\includegraphics[width=1\linewidth]{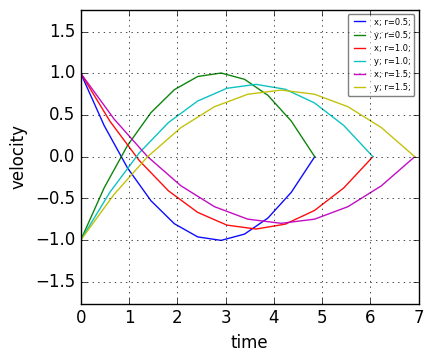}
		\caption{Comparison of velocity vs. time with different input weight}
		\label{trajectory_vel}
	\end{subfigure}
		\begin{subfigure}{0.4\textwidth}
		\centering
		\includegraphics[width=1\linewidth]{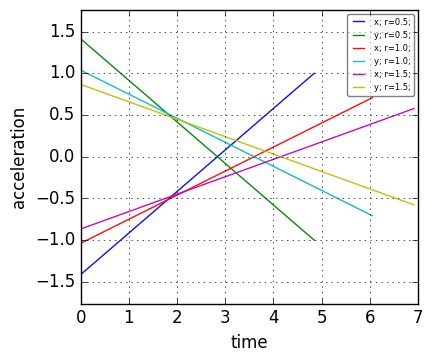}
		\caption{Comparison of acceleration vs. time with different input weight}
		\label{trajectory_acc}
	\end{subfigure}
	\caption{Comparison of Double Integrator Trajectory with different input weight}
	\label{trajectory1}
\end{figure}

We performed numerical simulation to verify the fixed-state free-final-time control. We compared the trajectories of the model across different value of input weight, $r$, from similar value initial and final state. We also compared the maximum control input among the trajectories of the double integrator model from the initial state $x_i=[2.3, -2.3, 1.0, -1.0]^{T}$ to the final state $x_f=[0.0, 0.0, 0.0, 0.0]^{T}$ on different input weight, $r$, with $r=1.5, \, r=1.0,$ and $r=0.5$. Figure \ref{trajectory1} shows the numerical experiment results of translational trajectories of the double integrator model.

%Figure \ref{trajectory1} and \ref{trajectory2} show the translational trajectories of the double integrator model and constrained angular trajectory, respectively. The translational trajectory was simulated from the initial state $x_i=[2.3, -2.3, 1.0, -1.0]^{T}$ to the final state $x_f=[0.0, 0.0, 0.0, 0.0]^{T}$ on different input weight, $r$, with $r=1.5, \, r=1.0,$ and $r=0.5$. The angular trajectory was simulated from the initial state of $\omega_i=-2.3\;rad,;\dot{\omega}_i=1.0\;rad/s$ to final state of $\omega_f=0.0\;rad$. 

Figure \ref{trajectory_pos} shows the numerical experiment results of the position of the model in $xy$-plane versus time, the position was measured in meters while the time was measured in seconds. Figure \ref{trajectory_vel} shows the numerical experiment results of the velocity of the model in $xy$-plane versus time, the position was measured in $m/s$ while the time was measured in seconds. Figure \ref{trajectory_acc} shows the numerical experiment results of the position of the model in $xy$-plane versus time, the position was measured in $m/s^2$ while the time was measured in seconds. 

\begin{table}[ht!]
\caption{Arrival Time and Maximum Control Input of Translational Trajectory for Several Input Weight}
\label{trajectories_table}
\centering
\begin{tabular}{|l|l|l|}
\hline
$r$ & arrival time (s) & maximum $\| u \|$ \\ \hline
$1.5$ & $6.9187936337$ & $1.2253000912634624$ \\ \hline
$1.0$ & $6.05276367644$ & $1.467295152420136$ \\ \hline
$0.5$ & $4.84707681233$ & $1.997746119057331$ \\ \hline
\end{tabular}
\end{table}
Table \ref{trajectories_table} shows the arrival time and the maximum control input for each input weights, where $\|u\| = \sqrt{u_x^2+u_y^2}$. It could be noted that penalizing more control input results on lower maximum acceleration and longer arrival time. Input weight, $r$, provides a 'soft' constraint for the translational trajectory.

\begin{figure}[ht!]
	\centering
	\begin{subfigure}{0.4\textwidth}
		\centering
		\includegraphics[width=1\linewidth]{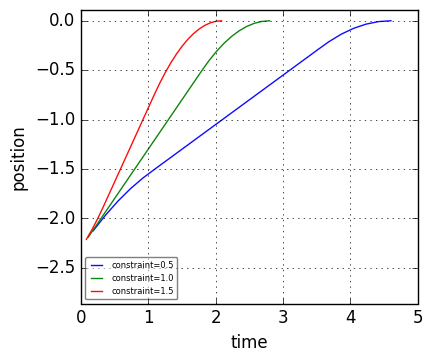}
		\caption{Comparison of heading angle vs. time with different constraint}
		\label{trajectory_pos_angular}
	\end{subfigure}
	\begin{subfigure}{0.4\textwidth}
		\centering
		\includegraphics[width=1\linewidth]{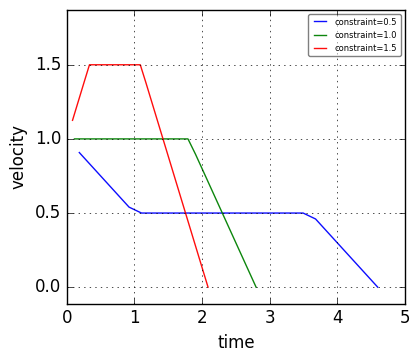}
		\caption{Comparison of angular velocity vs. time with different constraint}
		\label{trajectory_vel_angular}
	\end{subfigure}
		\begin{subfigure}{0.4\textwidth}
		\centering
		\includegraphics[width=1\linewidth]{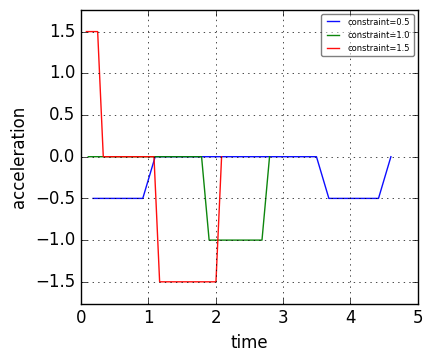}
		\caption{Comparison of angular acceleration vs. time with different constraint}
		\label{trajectory_acc_angular}
	\end{subfigure}
	\caption{Comparison of Angular Trajectory with different Constraints}
	\label{trajectory2}
\end{figure}

Figure \ref{trajectory_pos_angular} shows the numerical experiment results of the heading angle of the angular trajectory versus time, the angle was measured in radians while the time was measured in seconds. Figure \ref{trajectory_vel_angular} shows the numerical experiment results of the angular velocity of the angular trajectory versus time, the angle was measured in $rad/s$ while the time was measured in seconds. Figure \ref{trajectory_acc_angular} shows the numerical experiment results of the angular acceleration of the angular trajectory versus time, the angle was measured in $rad/s^2$ while the time was measured in seconds.

\begin{table}[ht!]
\caption{Arrival Time and Maximum Acceleration of Rotational Trajectory for Several Constraints}
\label{angular_trajectory_table}
\centering
\begin{tabular}{|l|l|l|}
\hline
Constraints                                                   $[\dot{\omega}_{max}\,\ddot{\omega}_{max}]$ & arrival time ($s$) & max acceleration \\ \hline
$[0.5\;rad/s;\;0.5\;rad/s^2]$ & $4.6$   & $0.5\,rad/s^2$           \\ \hline
$[1.0\;rad/s;\;1.0\;rad/s^2]$ & $2.8$   & $1.0\,rad/s^2$           \\ \hline
$[1.5\;rad/s;\;1.5\;rad/s^2]$ & $2.089$ & $1.5\,rad/s^2$         \\ \hline
\end{tabular}
\end{table}

We evaluated our numerical experiment on minimum-time trajectory for rotational motion of omnidirectional robot. We performed our numerical experiment on three cases with different constraint but with same initial and final state, $\omega_{i}=-2.3\;rad$, $\dot{\omega}_{i}=1.0\;rad/s$, and $\omega_{f}=0.0\;rad$. Figure \ref{trajectory2} show the numerical experiment results of the constrained angular trajectory.

Table \ref{angular_trajectory_table} shows the arrival time for each experiments. It could be noted that the maximum acceleration of rotational motion satisfying 'hard' constraint of the motion.

\begin{figure}[ht!]
	\centering
	\includegraphics[width=0.4\textwidth]{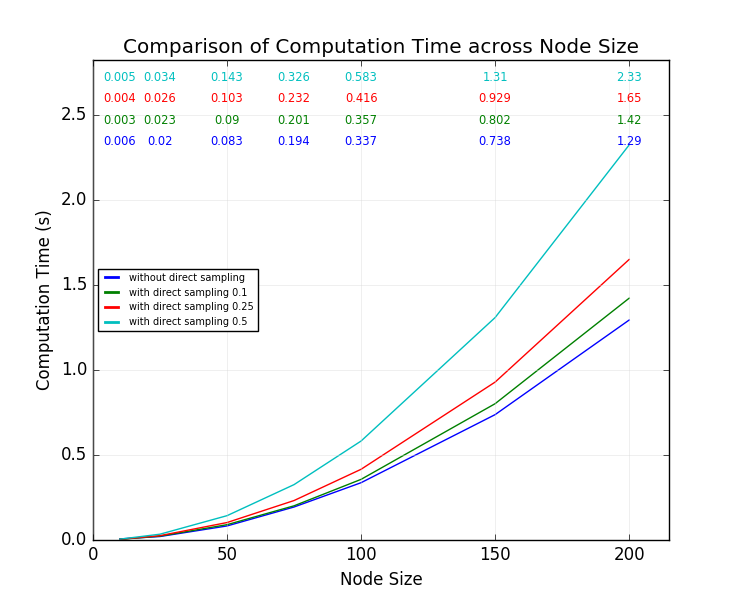}
	\caption{The Computation Time to grow the Tree}
	\label{plot_compare_all:time}
\end{figure}
\begin{figure}[ht!]
	\centering
	\includegraphics[width=0.4\textwidth]{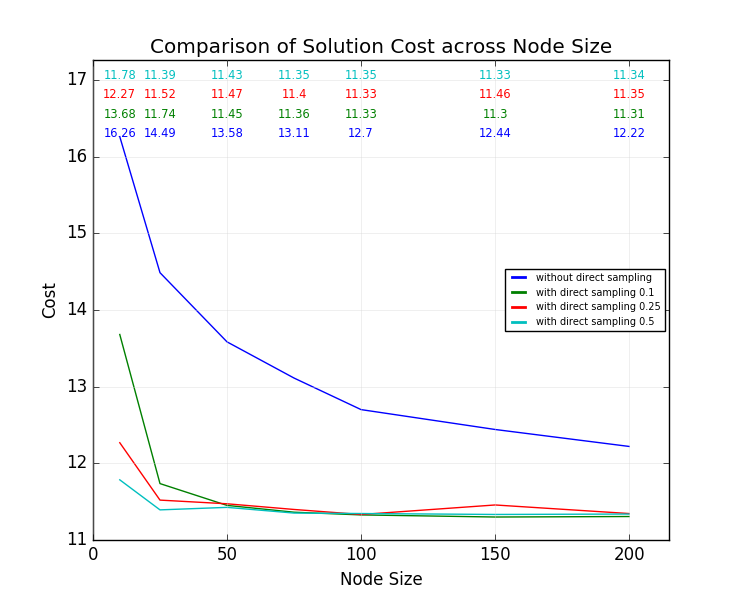}
	\caption{The Solution Cost of the Tree}
	\label{plot_compare_all:cost}
\end{figure}
\begin{figure}[ht!]
	\centering
	\begin{subfigure}{0.4\textwidth}
		\centering
		\includegraphics[width=1\linewidth]{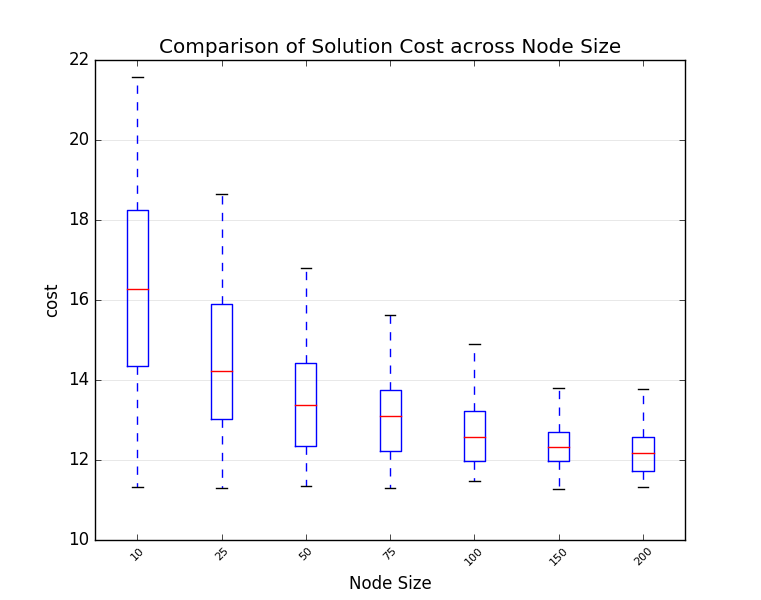}
		\caption{Solution Cost without Direct Sampling}
		\label{plot_compare_cost:no}
	\end{subfigure}
	\begin{subfigure}{0.4\textwidth}
		\centering
		\includegraphics[width=1\linewidth]{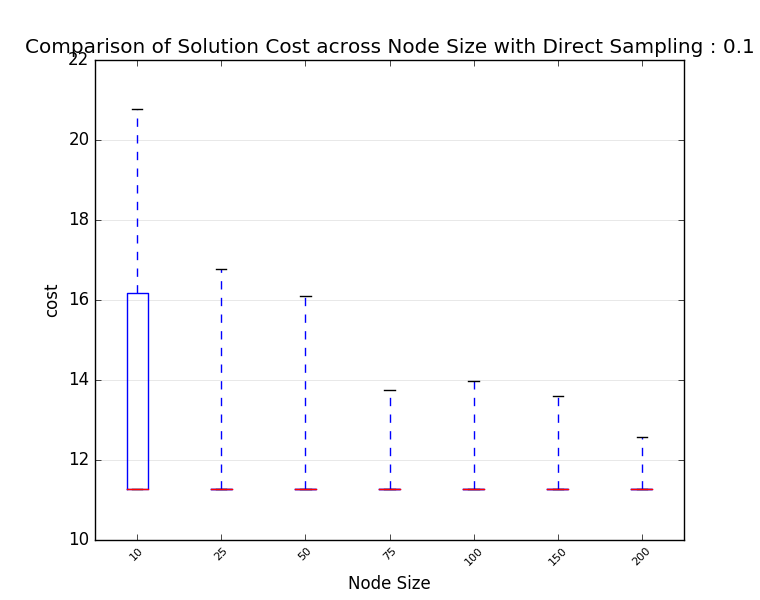}
		\caption{Solution Cost without Direct Sampling with $p=0.1$}
		\label{plot_compare_cost:0.1}
	\end{subfigure}
	\caption{Comparison of Solution Cost with and without Direct Sampling}
	\label{plot_compare_cost}
\end{figure}

\subsection{Performance Analysis}

\subsubsection{Computation Time}
We compare the computation time of the motion planner to find a solution. The comparison was done on different size of node grown and different direct sampling parameter. Figure \ref{plot_compare_all:time} shows the comparison of the mean computation time needed to grow the tree to a certain node size. 

It is shown that the computation time grows exponentially as the node size increases. It is also shown that by directly sampling the goal state, the computation time also increases. While this effect is small on lower number of nodes, it becomes more evident on larger number of nodes. 

On lower number of nodes, the difference of computation time between different direct sampling is relatively small compared to that with larger number of nodes. This is mainly because the planner too frequently tries to connect the goal state and less exploring the other possible state. This effect could be caused by the inserted goal state already has the lowest possible cost but the sampler keeps returning the goal state, hence much of computation time wasted while it could not grow the tree.

\subsubsection{Solution Cost}
Figure \ref{plot_compare_all:cost} shows the comparison of the mean solution cost of the growth tree across several node size. It is shown that by directly sampling the goal state reduce the cost of the solution trajectory. While it greatly reduce the cost on lower number of nodes, it should also be noted that more frequent direct sampling does not further reduce the cost on higher number of nodes. 

Figure \ref{plot_compare_cost} shows the data of solution costs from $100$ times of experiment with and without direct sampling. Figure \ref{plot_compare_cost:no} shows the solution cost without directly sampling the goal state and figure \ref{plot_compare_cost:0.1} shows the distribution of the solution cost with direct sampling to the goal with $0.1$ probability. 

Without direct sampling to goal state, the median of solution costs becomes smaller as the number of node increases, while with direct sampling to the goal state, the median of the costs is already on the lower range of solution costs even with smaller number of nodes and further increase the number of nodes does not reduces the median of solution costs. On both cases, however, increasing the number of nodes tends to reduce the range of maximum costs.

\urlstyle{tt}

\subsection{Dynamic Simulation Experiment}
We performed dynamic simulation using ROS-based RoboCup MSL simulation framework as proposed in \cite{yao2015simulation}. Table \ref{online_exp_settings} shows the configuration for the experiment. In this experiment, the controlled robot is commanded to move from one corner of the field to another corner without colliding with another robots. While another robots act as obstacles ignoring other robots' motions. Our implementation of the motion planning and trajectory tracking system is available at \url{https://github.com/alifahrri/robosoccer_motion_planning}. Video demonstration of this experiment is available at \url{https://youtu.be/jYTKOSrrcoY}. 

\begin{table}[ht!]
\caption{Experiment Settings}
\label{online_exp_settings}
\centering
\begin{tabular}{|l|l|}
\hline
Parameter                                      & Value                 \\ \hline
target tree size                               & $75$                  \\ \hline
direct sampling probability, $p$               & $0.1$                 \\ \hline
input weight, $r$                              & $0.5$                 \\ \hline
angular trajectory constraint, $[\dot{\omega}_{max},\,\ddot{\omega}_{max}]^T$ & $[0.75,\,0.75]^T$     \\ \hline
$a_I = [a_{I1},\,a_{I2},\,a_{I3}]^T$                 & $[5.0,\,5.0,\,5.0]^T$ \\ \hline
$a_P = [a_{P1},\,a_{P2},\,a_{P3}]^T$                 & $[3.0,\,3.0,\,3.0]^T$ \\ \hline
\vars{f sim}								& $60$ Hz                 \\ \hline
\vars{f plan}								& $3$ Hz                 \\ \hline
\vars{f control}							& $60$ Hz                 \\ \hline
\end{tabular}
\end{table}

\begin{figure*}[ht!]
	\begin{subfigure}{.4\textwidth}
		\centering
		\includegraphics[width=1\linewidth]{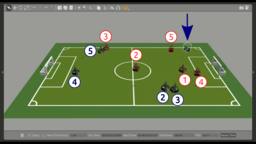}
		\caption{$t=423.5\;s$}
		\label{exp_out:out2}
	\end{subfigure}
	\begin{subfigure}{.4\textwidth}
		\centering
		\includegraphics[width=1\linewidth]{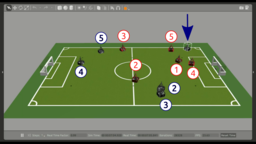}
		\caption{$t=424.9\;s$}
		\label{exp_out:out3}
	\end{subfigure}
	\begin{subfigure}{.4\textwidth}
		\centering
		\includegraphics[width=1\linewidth]{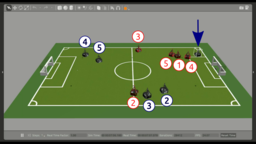}
		\caption{$t=426.2\;s$}
		\label{exp_out:out4}
	\end{subfigure}
	\begin{subfigure}{.4\textwidth}
		\centering
		\includegraphics[width=1\linewidth]{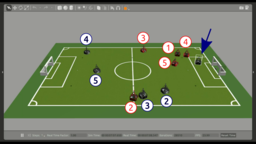}
		\caption{$t=427.6\;s$}
		\label{exp_out:out5}
	\end{subfigure}
	\begin{subfigure}{.4\textwidth}
		\centering
		\includegraphics[width=1\linewidth]{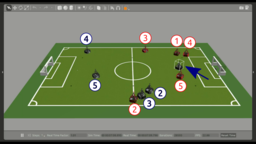}
		\caption{$t=429\;s$}
		\label{exp_out:out6}
	\end{subfigure}
	\begin{subfigure}{.4\textwidth}
		\centering
		\includegraphics[width=1\linewidth]{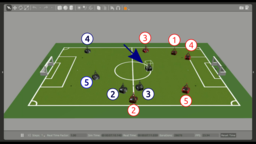}
		\caption{$t=430.4\;s$}
		\label{exp_out:out7}
	\end{subfigure}
	\begin{subfigure}{.4\textwidth}
		\centering
		\includegraphics[width=1\linewidth]{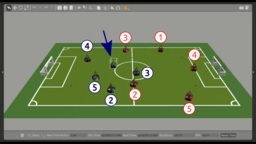}
		\caption{$t=431.8\;s$}
		\label{exp_out:out8}
	\end{subfigure}
	\begin{subfigure}{.4\textwidth}
		\centering
		\includegraphics[width=1\linewidth]{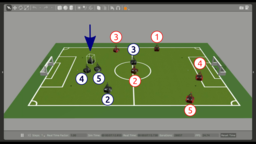}
		\caption{$t=433.2\;s$}
		\label{exp_out:out9}
	\end{subfigure}
	\begin{subfigure}{.4\textwidth}
		\centering
		\includegraphics[width=1\linewidth]{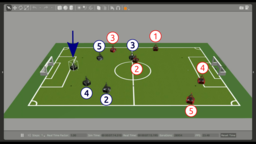}
		\caption{$t=434.6\;s$}
		\label{exp_out:out10}
	\end{subfigure}
	\begin{subfigure}{.4\textwidth}
		\centering
		\includegraphics[width=1\linewidth]{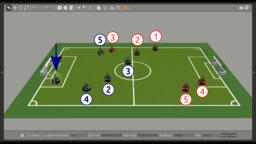}
		\caption{$t=436\;s$}
		\label{exp_out:out11}
	\end{subfigure}	
	\centering
	\caption{Snapshots of Online Motion Planning Experiment on Gazebo}
	\label{exp_out}
\end{figure*}

\begin{figure*}[ht!]
	\begin{subfigure}{.4\textwidth}
		\centering
		\includegraphics[width=1\linewidth]{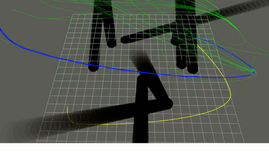}
		\caption{$t=423.5\;s$}
		\label{exp_outviz:out2}
	\end{subfigure}
	\quad
	\begin{subfigure}{.4\textwidth}
		\centering
		\includegraphics[width=1\linewidth]{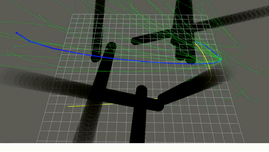}
		\caption{$t=424.9\;s$}
		\label{exp_outviz:out3}
	\end{subfigure}
	\begin{subfigure}{.4\textwidth}
		\centering
		\includegraphics[width=1\linewidth]{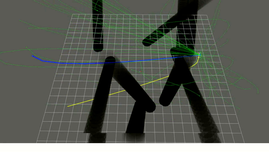}
		\caption{$t=426.2\;s$}
		\label{exp_outviz:out4}
	\end{subfigure}
	\quad
	\begin{subfigure}{.4\textwidth}
		\centering
		\includegraphics[width=1\linewidth]{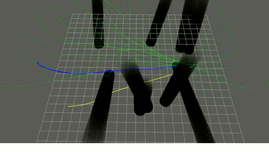}
		\caption{$t=427.6\;s$}
		\label{exp_outviz:out5}
	\end{subfigure}
	\begin{subfigure}{.4\textwidth}
		\centering
		\includegraphics[width=1\linewidth]{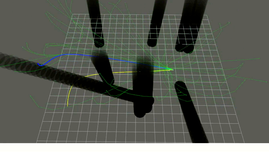}
		\caption{$t=429\;s$}
		\label{exp_outviz:out6}
	\end{subfigure}
	\quad
	\begin{subfigure}{.4\textwidth}
		\centering
		\includegraphics[width=1\linewidth]{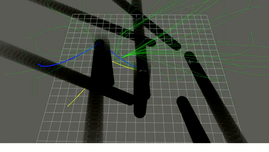}
		\caption{$t=430.4\;s$}
		\label{exp_outviz:out7}
	\end{subfigure}
	\begin{subfigure}{.4\textwidth}
		\centering
		\includegraphics[width=1\linewidth]{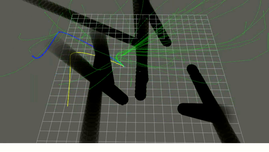}
		\caption{$t=431.8\;s$}
		\label{exp_outviz:out8}
	\end{subfigure}
	\quad
	\begin{subfigure}{.4\textwidth}
		\centering
		\includegraphics[width=1\linewidth]{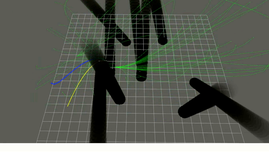}
		\caption{$t=433.2\;s$}
		\label{exp_outviz:out9}
	\end{subfigure} 
	\begin{subfigure}{.4\textwidth}
		\centering
		\includegraphics[width=1\linewidth]{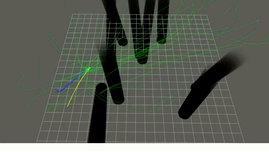}
		\caption{$t=434.6\;s$}
		\label{exp_outviz:out10}
	\end{subfigure} 
	\quad
	\begin{subfigure}{.4\textwidth}
		\centering
		\includegraphics[width=1\linewidth]{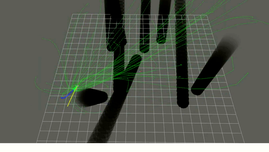}
		\caption{$t=436\;s$}
		\label{exp_outviz:out11}
	\end{subfigure}	
	\centering
	\caption{Snapshots of Kinodynamic-RRT* Visualization on RViz}
	\label{exp_outviz}
\end{figure*}

Figure \ref{exp_out} shows the snapshots of the online motion planning experiment on Gazebo simulator. The controlled robot is marked with white bounding box. Figure \ref{exp_outviz} shows the snapshots of the 3D visualization of kinodynamic-RRT* algorithm in RViz. In the figure, blue lines represent the planned trajectory in time-space with $z$-direction illustrate the arrival time, while yellow lines are the projections of the planned trajectory in $xy$-plane, and green lines illustrate the explored trajectory. The obstacles are visualized with black circles. Both of these snapshots are also taken with $1.4$ s interval.

\begin{figure}[ht!]
	\begin{subfigure}{0.45\textwidth}
		\centering
		\includegraphics[width=1\linewidth]{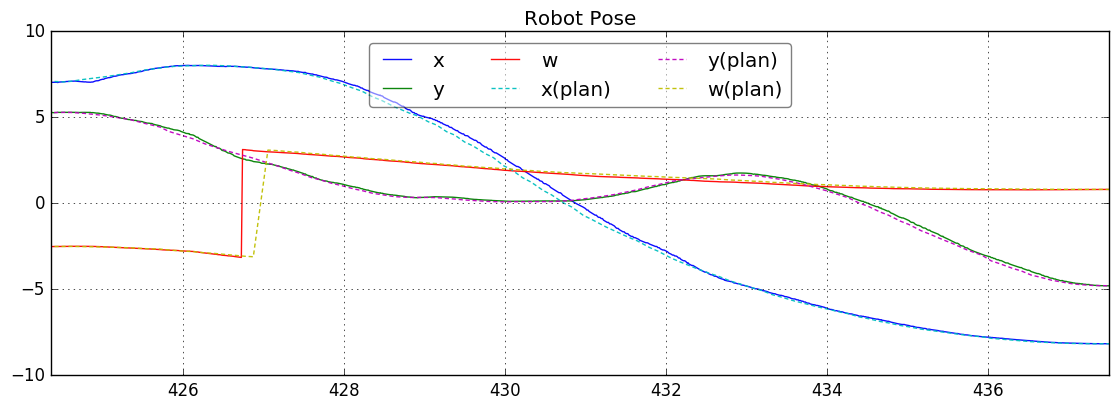}
		\caption{Planned and Realized Robot Pose versus Time}
		\label{exp_plot:pos}
	\end{subfigure}
	\begin{subfigure}{0.45\textwidth}
		\centering
		\includegraphics[width=1\linewidth]{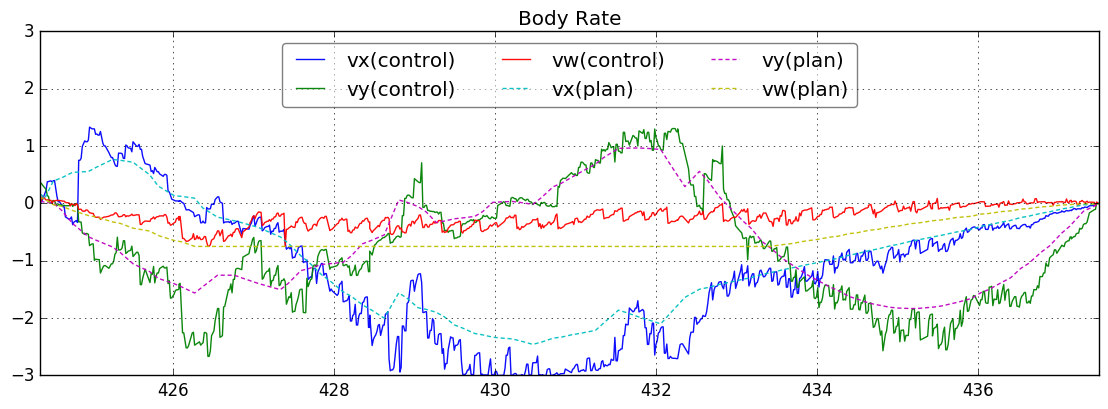}
		\caption{Planned and Realized Robot Velocity versus Time}
		\label{exp_plot:vel}
	\end{subfigure}	
	\centering
	\caption{Planned and Realized Trajectory}
	\label{exp_plot}
\end{figure}

Figure \ref{exp_plot} shows the planned and realized trajectory of the controlled robot. Figure \ref{exp_plot:pos} shows the robot pose over time with the realized poses are shown with solid lines while planned poses are shown with dashed lines. Figure \ref{exp_plot:vel} shows the robot's velocity over time with the realized velocities are shown in solid lines and planned velocities are shown in dashed lines. It is shown that the trajectory tracker is able to closely follows the planned pose from the motion planner, while the body rate roughly follows the planned velocity to track the reference trajectory. 

By taking the obstacles motion into account, the system could anticipate its movement in short time span resulting more reliable trajectory plan even if the motion planning loop rate is far lower than the simulator and trajectory tracking loop rate.

\section{Conclusions and Future Works} \label{conclusion}

\emph{Kinodynamic motion planning} for omnidirectional mobile robot was presented in this work. Obstacles' motion and robot's dynamic was taken into account while planning the collision-free trajectory. Dynamic constraints for translational motion was addressed by 'soft' constraint via input control weight while 'hard' constraints for rotational motion was satisfied by setting maximum allowed velocity and acceleration. The presented sampling strategy was able to reduce solution cost. Online planning \& tracking scheme was presented. Dynamic simulation shows the scheme was successfully applied to the simulated robot and environment.

There are several other problems that emerge during this work that should be tackled in the future research. For example,  \emph{synchronizing} the motion in translation and rotation by taking the robot's motion in body frame into account. The trajectory tracking controller only considers the current reference pose to compute the control input, while the subsequent pose is already known. Hence, we suggest future works should take this subsequent reference pose into account, e.g. using Model Predictive Controller. The proposed motion planning strategy assuming that the robot's and obstacles' states are perfectly known and accurate, while in real world this may be partially known and uncertain. So, we suggest to take this uncertainty into account while solving the trajectory, e.g. motion planning under uncertainty. To increase the quality of solution trajectory as well as computation efficiency it is recommended to develop and implement heuristic for the sampling procedure.

\bibliography{reference}

\end{document}